\documentclass[letterpaper,twocolumn,10pt]{article}
\usepackage{usenix2019_v3}

\usepackage{tikz}
\usepackage{amsmath}
\usepackage{subcaption}
\usepackage{pgfplots}
\usepackage{makecell}
\usepackage{filecontents}
\usepackage[inline]{enumitem}
\usepackage{pgfplotstable}
\usepackage{hyperref}
\usepackage{fancyhdr}

\pgfplotsset{compat=1.11,
        /pgfplots/ybar legend/.style={
        /pgfplots/legend image code/.code={%
        \draw[##1,/tikz/.cd,bar width=3pt,yshift=-0.2em,bar shift=0pt]
                plot coordinates {(0cm,0.8em)};},
},
}

\begin{document}

\date{}

\title{\Large \bf Hazard Detection in Supermarkets using Deep Learning on the Edge}

 \author{
 {\rm M.~G.~Sarwar Murshed}\\
 Clarkson University
 \and
 {\rm Edward Verenich}\\
 Air Force Research Laboratory\\
 Clarkson University
 \and
 {\rm James~J.~Carroll}\\
 Clarkson University
 \and
 {\rm Nazar Khan}\\
 Punjab University\\ College of Information Technology    
 \and
 {\rm Faraz Hussain}\\
 Clarkson University
} 

\maketitle

\begin{abstract}
Supermarkets need to ensure clean and safe environments for both shoppers and employees.
Slips, trips, and falls can result in injuries that have a physical as well as financial cost.
Timely detection of hazardous conditions such as spilled liquids
or fallen items on supermarket floors can reduce the chances of serious injuries.
This paper presents EdgeLite, a novel, lightweight deep learning model for
easy deployment and inference on resource-constrained devices.
We describe the use of EdgeLite on two edge devices for detecting supermarket floor hazards.
On a hazard detection dataset that we developed,
EdgeLite,  when deployed on edge devices, 
 outperformed six state-of-the art object detection models in terms of accuracy
while having comparable memory usage and inference time.
\end{abstract}

\section{Introduction}
In recent years, deep learning (DL) techniques have been widely applied in
a variety of domains (e.g. for  mobile robots and autonomous vehicles)
for making real-time decisions based on the surrounding environment \cite{BRUNETTI201817}.
However, deep learning models need a lot of computational power,
a sufficiently large memory, and dedicated hardware to function at a reasonable rate.
These constraints have traditionally inhibited the deployment of DL models
on resource-scarce devices.

With the improvement in edge computing technology,
many edge devices, such as the Coral Accelerator\footnote{The Coral USB Accelerator (\url{https://coral.ai/products/accelerator/}) can perform 4 trillion operations per second (TOPS) using only $0.5$ watts.}  
now  have significantly better computational and storage capabilities.
Therefore, edge computing is rapidly becoming a pervasive
technology for deploying deep learning models at the network edge.
Low-power devices such as the Raspberry Pi, the Coral Dev Board, the Intel Movidius Neural Compute Stick, and
BeagleBone AI are among the popular edge devices being used for for reducing automation latency.

\subsection*{Supermarket automation} There is increasing interest from industry,
especially from supermarkets,
to do repetitive \& dangerous work using robots instead of humans.
Moreover, the number of supermarkets and their customers is increasing every year.
According to a recent survey, there are currently more than 38,000 supermarkets in the United States \cite{statista} and on an average day, more than 32 million Americans visit grocery stores to pick up the daily necessities \cite{StartupPort}.

A large number of people are vulnerable to 
supermarket hazards on a daily basis.
Supermarket owners have to appoint permanent cleaners to keep the store clean and safe for shoppers
and employees.
Despite that, accidents are common. One study reports that slips, trips, falls and contact with objects and equipment has increased to 17.3 per 10,000 workers in 2014 from 12.1 in 2009 \cite{JoeBushSafty}.
Grocery  stores often need to pay large amounts of money when customers or workers
are injured in accidents e.g. a large retailer was ordered to pay more than \$400,000 to a customer
who slipped on a puddle of liquid soap \cite{GoguenAcc}.
Thus, maintaining a clean environment across the grocery store
is very important for the viability and profitability of businesses.

In the retail industry, efforts are increasingly been made to reduce human involvement in
hazard detection and inventory maintenance. 
Robots are being increasingly used for automated hazard detection
 and most of this technology involves using cloud servers for data processing \cite{shahrobotics}.
Transferring raw data to cloud servers increases communication costs,
causes delayed system response, and makes any private data vulnerable to compromise.
It is therefore desirable for the data to be processed as close to the source as possible.

\subsection*{Research contribution}
This paper describes a new technique for deploying more intelligence to the network edge,
allowing resource-constrained devices to aid faster decision-making. 
We have developed a model that processes supermarket image data on edge devices,
in order to detect hazardous floor conditions.
The early detection of  hazards such as spills and debris,
can prevent potentially serious accidents.

A major barrier to deploying deep learning models on edge devices or autonomous robots,
is the lack of efficient algorithmic systems to classify hazard images that are robust enough
to work with the limited computational resources and low battery life.
Moreover, there is no publicly available dataset for supermarket hazards.
To address these problems, this paper makes the following contributions:

\begin{itemize}
\item a new dataset of supermarket hazards images
\item the design of \emph{EdgeLite}, a lightweight image recognition CNN architecture
for detecting the presence or absence of supermarket floor hazards
\item a thorough comparison of EdgeLite with six state-of-the art deep learning
models (viz MobileNetV1, MobileNetV2, InceptionNet
V1, InceptionNet V2, ResNet V1, GoogleNet) for supermarket hazard detection when deployed on edge devices,
which shows EdgeLite to have the highest accuracy and comparable resource usage.
\end{itemize}

The rest of the paper is organized as follows:  Section \ref{sec:bkgroun}
presents prior work on hazard detection using deep learning and edge computing.
Section \ref{sec:ourapproach} gives the design and architecture of EdgeLite.
Experiments and performance evaluation of EdgeLite are described in Section \ref{sec:expRes},
which is followed by the  conclusion. 
        
\section{Related Work} \label{sec:bkgroun}
Despite the challenge posed by the computational requirements of deep learning,
several researchers have explored ways of deploying state-of-the-art DL systems in resource-constrained settings. For example, Hsu et al. developed a fall hazard detection system that generates an alert message when an object falls \cite{Hsu7988590}. 
Their approach has three key phases: 
\begin{enumerate*}
\item a skeleton extraction performed for building an ML prediction model to detect falls,
\item a Raspberry Pi 2 is used as an edge computing device for primary data processing and to reduce the size of videos/images, 
\item  and finally, falls are detected using machine learning inference on the cloud, and users are notified in appropriate cases.
\end{enumerate*}

A hierarchical distributed computing architecture has been designed for a smart city to analyze big data at the edge of the network to detect hazardous events in a city area \cite{Tang7874167}. A working prototype was constructed using a machine learning algorithm on low-power computing devices to quickly detect hazardous events to avoid potential damages. Researchers have also used deep learning models to detect small-sized hazards on roads (e.g. lost cargo) which is a vital capability for autonomous vehicles \cite{Ramos7995849}.
A food image recognition system for automatic dietary assessment, introduced by Liu et al \cite{Liu2018foodrec}, used  deep learning models deployed
collaboratively on edge devices and cloud servers.

Most research on hazard detection has used both edge and cloud computing simultaneously to identify specific events.
Such an approach cannot be easily used if it has to work using edge
devices only. 
Moreover, if there multiple distinct classes but with several common features,
the complexity of the deep learning model makes its deployment on resource-contrained devices more challenging.
Therefore, we have conducted extensive experiments to measure the accuracy of several state-of-the-art
CNN models, as well as EdgeLite, on edge devices.

\section{Approach} \label{sec:ourapproach}
Fitting CNNs on edge devices is challenging due to their limited memory and computer power.
For example, an edge device may not have enough memory to store parameters, the weight values of CNN filters,
or the input data arrays.
Therefore, there is increasing interest in lightweight, compute-efficient CNNs.
We tested several pre-trained models (\autoref{table:memacc}), using transfer learning to train them on our
supermaket hazard dataset, and then developed a new architecture \emph{EdgeLite}
that outperforms other models for hazard detection on resource-constrained devices.

There are two main approaches used to fit CNNs on edge devices. The first is by reducing
the number of mathematical operations required for a model with minimal accuracy losses and improving inference time.
Examples of this category are MobileNet \cite{Howard17MNet}, SqueezeNet \cite{Iandola16SqueezeNet},
EfficientNet \cite{tan2019efficientnet}, and 
ShuffleNet \cite{Zhang18ShuffleNet}.
The second approach is to quantize the model weights from higher bit floating point (e.g. 32 bit) into lower
bit-depth representations (e.g. 8 bit). This technique is exemplified by Binary Neural Networks
(BNN) \cite{hubara2016binarized} and XNOR-Net \cite{Rastegari2016XNORNetIC}.

This paper describes EdgeLite, a simplified CNN model that requires fewer mathematical operations.
We quantized this model to run on resource-constrained edge devices.
We experimented with a varying number  of layers in order to come up with a model that can run
on resource-contrained devices (such as the Coral Dev Board and the Raspberry Pi) while retaining high accuracy.
We trained our model on ImageNet, replaced the final layer of the neural network
with a binary classifier and then used transfer-learning to fine-tune the model on our own hazard dataset.

\subsection{EdgeLite Architecture}
In order to have a model that can be used for inference on resource-constrained edge devices, 
with low memory footprint without any additional hardware, 
we developed a lightweight CNN architecture which achieved higher than 90\% accuracy on
our hazard dataset.

Our CNN architecture has 19 layers, not counting the pooling layers.
In order to extract features at different scales, we used filters with multiple sizes that operate on the same level.
The different types of filters used were of size $1 \times 1$, $3 \times 3$ and $5 \times 5$. 
To make the network computationally cheaper,  $1 \times 1$ convolutions were used to reduce
the input channel depth and an extra $1 \times 1$ convolution was used before the
$3 \times 3$ and $5 \times 5 $ convolutions.

The vanishing gradient problem is one of the major issues of such deep learning classifiers.
To prevent this, Szegedy et al. introduced two auxiliary layers to the middle of the network which prevent the middle part of the network from dying out, and also have a regularizing effect.
In EdgeLite, we used two auxiliary layer to solve the vanishing gradient problem; 
These layers are only used during training and they are discarded during inference.
Thus, the deployed model is not burdened by these extra layers.

Our CNN architecture is shown in \autoref{table:en_v1}, which consists of convolution,
max-pooling, avg-pooling and EdgeLite layers. 
EdgeLite layers are incorporated into CNNs as a way of reducing computational expense through a dimensionality reduction with stacked $1 \times 1$ convolutions.
Multiple kernel filter sizes are used in this layer and an extra $1 \time 1$
convolution is added  whenever $3 \times 3$ and $5 \times 5$ layers are used.
All the kernels are ordered to operate on the same level sequentially.
A max-pooling is performed in this layer and the resulting outputs are concatenated,
and then sent to the next layer.  
EdgeLite's architecture is inspired by Inception \cite{Szegedy2014GoingDW}.
However, we reduced the number of layers and size of the kernels to make it
suitable for resource-constrained devices.  
 
\begin{table}
\small
  \caption{The outline of the proposed EdgeLite architecture.}
  \centering
  \begin{tabular}{|c|{c}|c|}
    \hline
Type & Patch size/stride & Output size \\
\hline
Conv  & $7 \times 7 /2$  & $ 112 \times 112 \times 64 $ \\
\hline
Max pool  & $3 \times 3 / 2$  & $56 \times 56 \times 64 $ \\
\hline
Conv  & $3 \times 3 / 1$  & $56 \times 56 \times 192 $ \\
\hline
Conv  & $3 \times 3 / 1$  & $56 \times 56 \times 256 $ \\
\hline
Conv  & $3 \times 3 /1 $  & $56 \times 56 \times 480 $ \\
\hline
Pool & $3 \times 3 /2$  & $14 \times 14 \times 480 $ \\
\hline
$5 \times$ EdgeLite\_conv  &   & $14 \times 14 \times 832 $ \\
\hline
Pool  & $3 \times 3 /2$  & $7 \times 7 \times 832 $ \\
\hline
$2 \times$ EdgeLite\_conv  &   & $7 \times 7 \times 1024 $ \\
\hline
Pool  & $7 \times 7 / 1$  & $1 \times 1 \times 1024 $ \\
\hline
Dropout  &   & $1 \times 1 \times 1024 $ \\
\hline
Linear & &$1 \times 1 \times 1000 $ \\
\hline
Softmax  & Classifier  & $1 \times 1 \times 2 $ \\
\hline
\end{tabular}
\label{table:en_v1} 
\end{table}

\section{Experiments} \label{sec:expRes}
We trained six widely used CNN-based image classification models on our supermarket hazard dataset using MXNet and then compared their performance with EdgeLite in terms of model accuracy,
execution time,  and memory usage.
When deployed on the Raspberry Pi and the Coral Dev Board, EdgeLite outperformed the other models in detecting hazards in supermarket floor images.

\subsection{Dataset}
Since we know of no publicly available dataset for supermarket hazards, 
we built an original real-world dataset of images showing hazards in supermarket floors.
This dataset contains supermarket images labeled either as having a hazardous floor or not.
Since we were able to collect only 1180 images,
we also added synthetic images to enrich our dataset. 

We generated an additional 300 images using a data synthesis method
that we implemented in a small tool. 
First, we collected the images of common grocery items for example bakery, bread, broken eggs, sauces and liquid spills. Then, cropped and resized these images and put them on clean floor images. This tool used clean floor images as background images and then cropped hazards were placed on that background.

\begin{table}
\small
  \caption{The distribution of images in our supermarket floor hazards dataset.}
  \centering
  \begin{tabular}{|c|{c}|c|c|}
    \hline
{Class} & Training & Validation & Testing \\
\hline
Hazardous floor  & 2224  & 526 & 500\\
\hline
Clean floor  & 2224  & 526 & 500\\
\hline
\end{tabular}
\label{table:dataset} 
\end{table}

We also used data augmentation methods, including horizontal flip, shift, zoom, and brightness change, to generate an additional 5020 images.
After data augmentation, we had 5500 images for training \& validation and 1000 images for testing. \autoref{table:dataset} shows images split among training, validation and testing set. 

\subsection{Hyperparameter Tuning}
Hyperparameter tuning is arguably the most important factor for improving performance of CNN models.
We tuned multiple hyperparameters for EdgeLite as well as the other architectures used our experiments.
We paid special attention to the momentum, learning rate, weight decay coefficients, dropout rates, and corruption bounds for various data augmentations: random scalings, input pixel dropout and random horizontal reflections.
We optimized these over a validation set of slightly more than 1,000 examples drawn from the training set.
We used a grid search and varied the values of these hyperparameters
and ran each network for 300 epochs on the hazard dataset.
Due to the small size of our training set, we conducted extensive tuning experiments
(batch size: from 8 to 128, learning rate: from 0.0005 to 0.1, momentum: from 0.0 to 0.9 , decay: 0.00001 to 0.0001)
and evaluated the model with the best-performing hyperparameter configuration on the test set.
Python Keras and MXNet were used for composing, training, and evaluating the models.

\begin{table*}[!t]
  \small
  \renewcommand{\arraystretch}{2}
  \caption{Comparison between the performance of existing models with EdgeLite based on 
    experiments on the Coral Dev Board (CB),  Raspberry Pi (RP) as well as a desktop machine (DM).}  
  \label{table:memacc}
  \centering
  \begin{tabular}{|c|c|c|c|c|c|c|c|c|c|c|}      
    \hline
    DL & Model & \multicolumn{3}{l|}{Inference time (sec)} & \multicolumn{3}{c|}{Avg. RAM
Usage (MB)} & \multicolumn{3}{c|}{Accuracy} \\
     \cline{3-11}
     {Model} & {size (MB)} & DM & CB & RP & DM & CB & RP & DM & CB & RP\\
    \hline
    InceptionV1 & 6.8 & 0.12 & 0.56 & 0.7127& 23.1  & 7 & 3.26 & 87.9\% & 86.31\% & 86.41\%\\
    \hline
    InceptionV2 & 10.2 & 0.17 & 0.79 & 0.7293 & 23.4 & 9 & 3.65 & 86.131\% & 85.11\%& 85.13\%\\
    \hline
    ResNet & 5.87 & 0.192 & 0.58 & 0.75 & 25.7 & 8.5 & 5.87 & 85.65\% & 84.7\%& 84.672\%\\
    \hline
    GoogleNet & 5.051 & 0.11 & 0.55 & 0.71 & 27.6 & 8 & 7.19 & 89.41\% & 88.1\%& 88.1\%\\
    \hline
    MobileNetV1 & 3.3 & 0.074 & 0.286 & 0.69 & 22.5 & 6 & 2.89 & 89.75\% & 87.87\%& 86.8\%\\
    \hline
    MobileNetV2 & 2.3 & 0.075 & 0.268 & 0.68 & 22.4 & 5 & 2.74 & 89.5\% & 87.022\%& 85.9\%\\
    \hline
    EdgeLite & 4.9 & 0.098 & 0.506 & 0.69 & 24.3 & 5.2 & 3.1 & 92.701\% & 92.37\%& 91.981\%\\
    \hline
  \end{tabular}
\end{table*}

\subsection{Implementation}
We conducted our tests on two edge devices and one desktop machine.
The desktop machine was used solely to train the models. We have divided our experiment into three parts: 
\begin{enumerate}
\item training the model on desktop machine
\item compress the model so that it can be deployed on resource-constrained edge devices
\item perform inference after deploying the model on  the edge devices. 
\end{enumerate}

Training was done on a desktop machine with 20 CPU\footnote{Intel(R) Xeon(R) CPU E5-2690 v2 @ 3.00GHz} cores
and with 64 GB RAM. 
Tensorflow and MXnet were used to build and quantize the networks.
We tested using the  model that performs best on the validation set. 
We got the best model by training the networks using the Adam optimizer with a momentum of 0.9, and batch size of 32. The learning rate was 0.002, with decay of 0.00004 on the model weights.
After training, we evaluated the performance of the model on the testing dataset.

The trained neural network was exported to ONNX (Open Neural Network Exchange) format,
converted to a TensorFlow Lite flatbuffer file, and finally converted to a  TensorFlow Lite model
for reducing storage and inference time.
This conversion reduced the model size by more than 65\% compared to the original model.
We used the edgetpu\_compiler \cite{EdgeCompiler}
to compile the trained model. 
Finally, we deployed the compiled model on both the Coral Dev Board and the Raspberry Pi.   
  
\subsection{Results}
For analyzing the performance of EdgeLite and other state-of-the-art CNN models on the edge devices,
we measured the accuracy, the inference time required to classify an image,
and also the amount of memory used during classification. 
\autoref{table:memacc} summarizes the results. 
EdgeLite outperformed all other models in terms of accuracy.
The fastest model is the MobileNetV2, which only took 0.268 seconds for inference on the
Coral Dev board and 0.68 sec on the Raspberry Pi.
EdgeLite took 0.506 sec for inference on the Coral Dev and 0.69 sec on the Raspberry Pi.

Compared to MobileNet,  EdgeLite took more time and used more memory for classification
because the latter has more filters than MobileNet and the number of convolution filters in each layer
of a CNN has a significant effect on inference time and memory usage.
Another factor that has a large impact on inference time and memory usage is the number of pixels of input images.
\autoref{table:memacc} illustrates the memory usage during inference while using $224 \times 224$ images.
MobileNet V2 used only 5 MB memory for classification on the Coral Dev board while
EdgeLite used slightly more RAM of 5.2MBs.

In terms of memory usage, the Raspberry Pi outperformed all other devices.
All models took less memory for inference on the Pi.
MobileNet~V2 only took 2.74 MB and MobileNet~V1 took 2.89 MB while EdgeLite took 3.1 MB during inference.

\section{Conclusion} \label{sec:conclusion}
Onboard data analysis is rapidly becoming one of the key focus areas for AI researchers.
However, modern deep learning models typically have millions of parameters,
making their deployment on low-memory devices challenging.
Edge computing devices that can run the TensorFlow framework, like the Coral Dev Board and the Raspberry Pi,
are drawing more attention of researchers
because of their computational capabilities while having a low power footprint.

In this paper, we have demonstrated the advantages and limitations
of six state-of-the-art deep learning based architectures for object detection (\autoref{table:memacc}).
We also introduced a fast, lightweight, deep learning model (EdgeLite) for easy deployment \& inference
on resource-constrained edge devices.
EdgeLite  outperformed the other six models for detecting hazards in supermarket floors in terms of accuracy.

\section*{Availability}
The code used to generate the results reported in this paper
is available via a public repository: \url{https://github.com/sarwarmurshed/supermarket_hazard_detection}

\bibliography{bibfile}{}
\bibliographystyle{unsrt}
\end{document}